\documentclass[10pt,twocolumn,letterpaper]{article}

\usepackage[pagenumbers]{cvpr} 

\usepackage{graphicx}
\usepackage{amsmath}
\usepackage{amssymb}
\usepackage{booktabs}

\usepackage[subrefformat=parens]{subcaption}
\usepackage{float}
\usepackage{afterpage}
\usepackage{comment}
\usepackage{caption}
\usepackage{longtable}
\usepackage{array}
\usepackage{multirow}
\usepackage[accsupp]{axessibility}  

\usepackage[pagebackref,breaklinks,colorlinks]{hyperref}

\DeclareMathOperator*{\argmin}{argmin}

\usepackage[capitalize]{cleveref}
\crefname{section}{Sec.}{Secs.}
\Crefname{section}{Section}{Sections}
\Crefname{table}{Table}{Tables}
\crefname{table}{Tab.}{Tabs.}
\Crefname{figure}{Figure}{Figures}
\crefname{figure}{Fig.}{Figs.}

\def\cvprPaperID{5167} 
\def\confName{CVPR}
\def\confYear{2023}

\begin{document}

\title{High-Res Facial Appearance Capture from Polarized Smartphone Images}

\author{
Dejan Azinovi{\'c}$^{1}$~~~
Olivier Maury$^2$~~
Christophe Hery$^2$~~
Matthias Nie{\ss}ner$^1$~~~
Justus Thies$^{3}$~~~
\vspace{0.4cm} \\ 
$^1$Technical University of Munich~~~
$^2$Meta Reality Labs~~~
$^3$Max Planck Institute for Intelligent Systems~~~
}


\newcommand{\todo}[1]{\textcolor{red}{TODO: #1}}
\newcommand{\TODO}[1]{\textcolor{red}{TODO: #1}}
\newcommand{\DA}[1]{\textcolor{cyan}{DA: #1}}
\newcommand{\JT}[1]{\textcolor{magenta}{JT: #1}}
\newcommand{\MATTHIAS}[1]{\textcolor{green}{\textbf{Matthias: #1}}}
\newcommand{\OM}[1]{\textcolor{blue}{OM: #1}}
\newcommand{\CH}[1]{\textcolor{blue}{CH: #1}}

\newcommand{\x}{\mathbf{x}} 	
\newcommand{\y}{\mathbf{y}} 	
\newcommand{\n}{\mathbf{n}} 	
\newcommand{\wo}{\mathbf{i}}	
\newcommand{\wi}{\mathbf{o}}	
\newcommand{\X}{\mathbf{X}} 	

\newcolumntype{L}[1]{>{\raggedright\let\newline\\\arraybackslash\hspace{0pt}}m{#1}}
\newcolumntype{C}[1]{>{\centering\let\newline\\\arraybackslash\hspace{0pt}}m{#1}}
\newcolumntype{R}[1]{>{\raggedleft\let\newline\\\arraybackslash\hspace{0pt}}m{#1}}

\renewcommand{\paragraph}[1]{\vspace{.05cm}\noindent\textbf{#1}}


\twocolumn[{%
\renewcommand\twocolumn[1][]{#1}%
\maketitle
\begin{center}
    \centering
    \captionsetup{type=figure}
    \vspace{-0.5cm}
     \includegraphics[width=\textwidth]{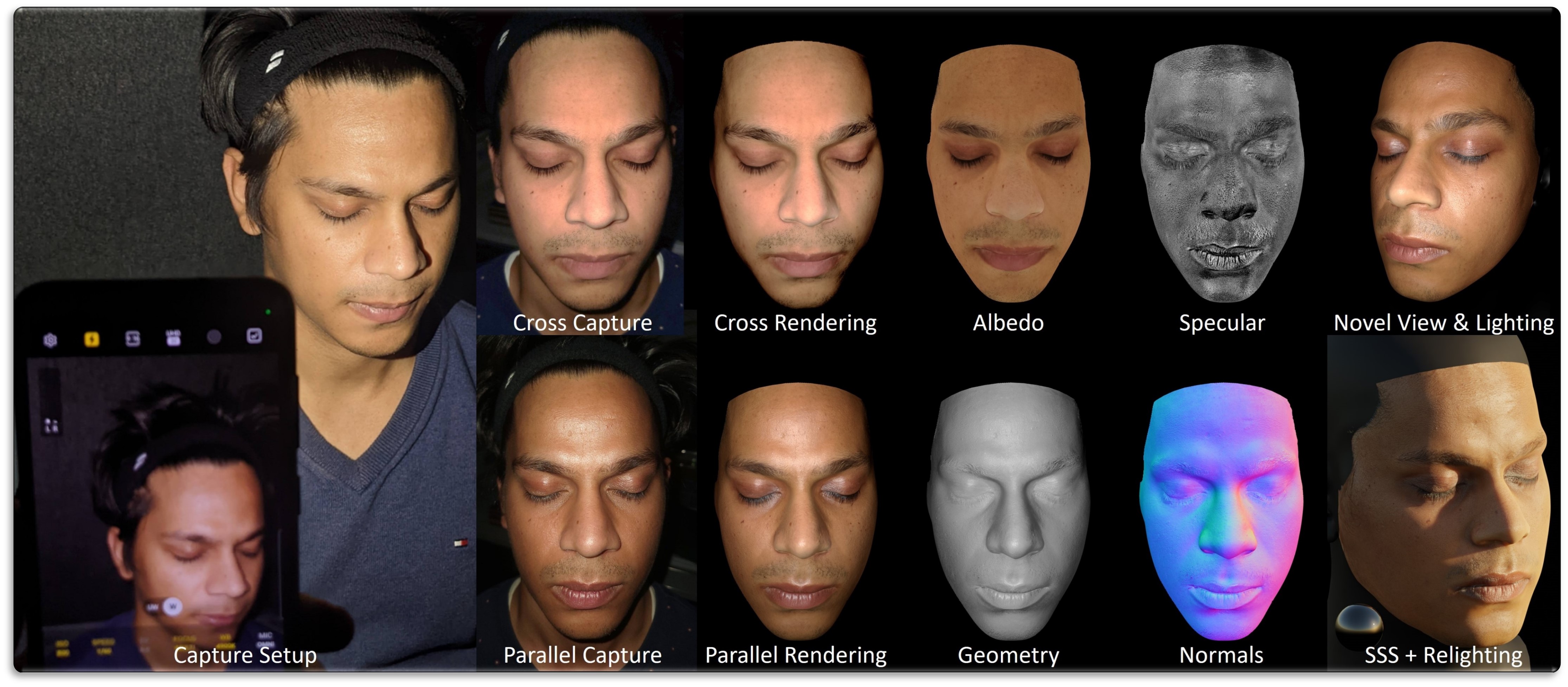}
      \caption{
        Our method obtains high-resolution skin textures from two RGB input sequences captured with polarization foils attached to a smartphone. The core idea is to separate the skin's diffuse and specular response by capturing one cross-polarized and one parallel-polarized sequence. We recover an accurate geometry with multi-view stereo, fit a parametric head model, and employ a differentiable rendering strategy to recover 4K diffuse albedo, specular gain and normal maps. These can be used with off-the-shelf rendering software, such as Blender, to produce photo-realistic images from novel views, under novel illumination and with subsurface scattering (SSS).
      }
      \label{fig:teaser}
\end{center}%
}]

\newcommand\blfootnote[1]{%
	\begingroup
	\renewcommand\thefootnote{}\footnote{#1}%
	\addtocounter{footnote}{-1}%
	\endgroup
}
\blfootnote{All data has been captured at the Technical University of Munich.}

\vspace{-0.3cm}
\begin{abstract}
We propose a novel method for high-quality facial texture reconstruction from RGB images using a novel capturing routine based on a single smartphone which we equip with an inexpensive polarization foil.
Specifically, we turn the flashlight into a polarized light source and add a polarization filter on top of the camera.
Leveraging this setup, we capture the face of a subject with cross-polarized and parallel-polarized light.
For each subject, we record two short sequences in a dark environment under flash illumination with different light polarization using the modified smartphone.
Based on these observations, we reconstruct an explicit surface mesh of the face using structure from motion.
We then exploit the camera and light co-location within a differentiable renderer to optimize the facial textures using an analysis-by-synthesis approach.
Our method optimizes for high-resolution normal textures, diffuse albedo, and specular albedo using a coarse-to-fine optimization scheme.
We show that the optimized textures can be used in a standard rendering pipeline to synthesize high-quality photo-realistic 3D digital humans in novel environments.
\end{abstract}

\section{Introduction}

In recent years, we have seen tremendous advances in the development of virtual and mixed reality devices.
At the same time, the commercial availability of such hardware has led to a massive interest in the creation of 'digital human' assets and photo-realistic renderings of human faces.
In particular, the democratization to commodity hardware would open up significant potential for asset creation in video games, other home entertainment applications, or immersive teleconferencing systems.
However, rendering a human face realistically in a virtual environment from arbitrary viewpoints with changing lighting conditions is an extremely difficult problem.
It involves an accurate reconstruction of the face geometry and skin textures, such as the diffuse albedo, specular gain, or skin roughness.
Traditionally, this problem has been approached by recording data in expensive and carefully calibrated light stage capture setups, under expert supervision.
We seek to simplify this capture process to allow individuals to reconstruct their own faces, while keeping the quality degradation compared to a light stage to a minimum.
The disentanglement of geometry and material of human faces is an extremely ill-posed problem.
Current solutions involve a capture setup with multiple cameras and light sources, with millimeter-accurate calibration.
A common approach to disentangling face skin surface from subsurface response is the use of polarization filters~\cite{debevec2000acquiring} in tandem with such expensive capture setups.
Given such a carefully calibrated capture setting, one can use differentiable rendering to estimate the individual skin parameters in an analysis-by-synthesis approach.
While these methods do produce visually impressive results, they are limited to high-budget production studios.

In this paper, we propose a capture setup consisting of only a smartphone and inexpensive polarization foils, which can be attached to the camera lens and flashlight.
Inspired by light stage capture setups, a user captures two sequences of their face, one with perpendicular filter alignment, and one with parallel alignment.
This allows for a two-stage optimization, where we first reconstruct a high-resolution diffuse albedo texture of a user's face from the cross-polarized capture, followed by recovery of the specular albedo, normal map, and roughness from the parallel-polarized views.
Data is captured in a dark room to avoid requiring pre-computation of an environment map.
In addition to visually compelling novel view synthesis and relighting results, our method produces editable textures and face geometry.

\medskip
\noindent
In summary, the key contributions of our project are:
\begin{itemize}
    \item We propose a commodity capture setup that combines a smartphone’s camera and flashlight with polarization foils. The polarization allows us to separate diffuse from specular parts, and to reconstruct the user’s face textures, such as diffuse albedo, specular albedo and normal maps.
    \item Our proposed capture setting with the co-located camera and light enables separation of skin properties from illumination, which is of key importance for realistic rendering of faces.
    \item We propose a coarse-to-fine optimization strategy with mip-mapping, which increases sharpness of the reconstructed appearance textures.
\end{itemize}
\begin{figure*}[t!]
    \begin{center}
    \includegraphics[width=\linewidth]{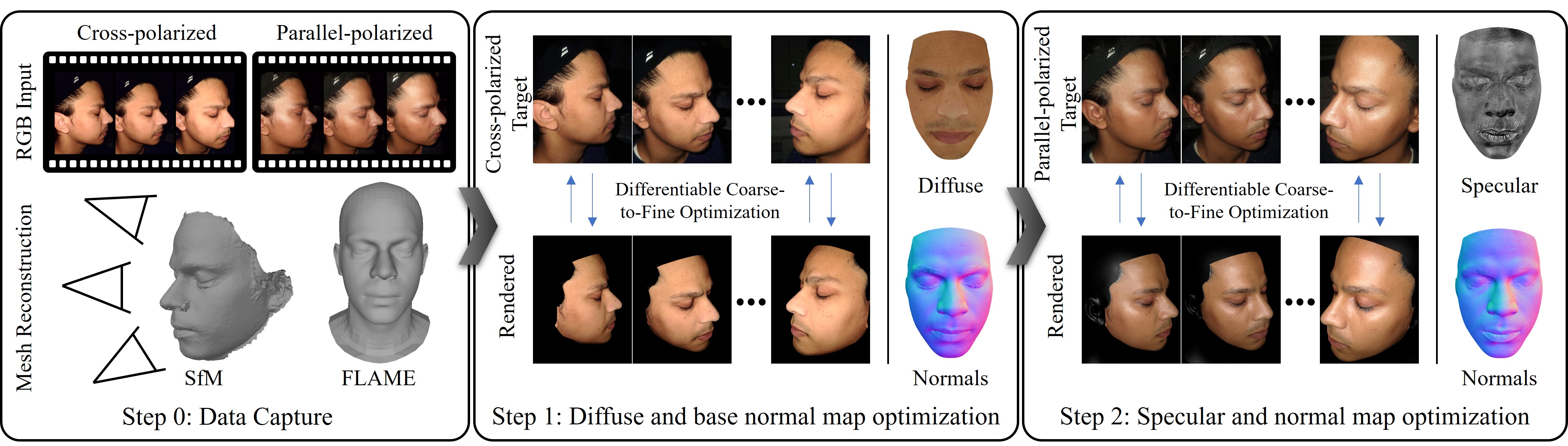}
    \end{center}
   \vspace{-0.2cm}
   \caption{
       Our optimization has three steps:
       In step 0, we capture data with a handheld smartphone which is equipped with polarization foils (on the camera, as well as on the flashlight; see Figure~\ref{fig:capture_setup}).
       We reconstruct the facial geometry and estimate camera poses based on all captured images using structure-from-motion and multi-view stereo.
       To ensure consistent texture parameterization across different subjects, we non-rigidly fit a FLAME mesh to the scan.
       In a subsequent photometric optimization step (step 1), we estimate a high-resolution diffuse texture of the skin from the cross-polarized data, as well as an initial normal map.
       The reconstructed geometry, diffuse and normal map are used as input for step 2 of the optimization.
       Using the parallel-polarized sequence, we estimate the specular gain and final normal map in a second photometric optimization. In addition, a global skin roughness value is optimized in this step.
    }
\vspace{-0.2cm}
\label{fig:pipeline}
\end{figure*}

\section{Related Work}

High-fidelity face appearance capture and reconstruction has received significant attention in the entertainment industry for creating digital humans and more recently in the AR/VR community for generating realistic avatars.
In our context, facial appearance reconstruction means recovering a set of high-resolution albedo, specular (gain and roughness) and normal maps.
Over the years, physically-based skin scattering models have become ever more sophisticated ~\cite{wrenninge2017path, chiang2016practical, klehm2015recent}; however, their input texture quality remains the single most important factor to photo-realism. 

\paragraph{Polarization.}
For some time, polarization has been used to separate specular from diffuse~\cite{muller1995polarization,wolff1993constraining, rahmann2001reconstruction}. These techniques rely on the fact that single bounce specular reflection does not alter the polarization state of incoming light.
Riviere et al.~\cite{riviere2017polarization} propose an approach to reconstruct reflectance in uncontrolled lighting, using the inherent polarization of natural illumination.
Nogue et al.~\cite{Nogue2022} recover SVBRDF maps of planar objects with near-field display illumination, exploiting Brewster angle properties.
Deschaintre et al.~\cite{Deschaintre21} use polarization to estimate the shape and SVBRDF of an object with normal, diffuse, specular, roughness and depth maps from a single view.
Dave et al.~\cite{dave2022pandora} propose a similar approach for multi-view data.
In MoRF~\cite{morf}, a studio setup with polarization is used to reconstruct relightable neural radiance fields of a face.

\paragraph{Lightstage capture systems.}
In their foundational work, Debevec et al.~\cite{debevec2000acquiring} introduced the Lightstage system to capture human face reflectance using a dome equipped with controlled lights, separating the diffuse from the specular component using polarization filters.
Follow-up work reconstructs high-resolution normal maps using photometric stereo~\cite{woodham1980photometric}, compensates for motion during the capture~\cite{wilson2010temporal} and expands the captured area~\cite{ghosh2011multiview}.

The proposed capture studios didn't come without limitations, as the lighting environment needed to be tightly controlled, the lighting patterns involved took a relatively long time, and the polarization filters were challenging to set up for multiple cameras and lights.
Fyffe et al.~\cite{fyffe2009cosine, fyffe2010single, fyffe2015single, fyffe2016near} proposed the use of color gradients and spectral multiplexing to reduce capture time.
With the objective of designing a more practical system, Kampouris et al.~\cite{kampouris2018diffuse} demonstrate that binary gradients are sufficient for separating diffuse from specular without polarization.
Lattas et al.~\cite{lattaspractical} use an array of monitors or tablets for a practical binary gradients capture studio.
In line with this thread of research, Gotardo et al.~\cite{disney_practical_dynamic} present a multi-view setup for dynamic facial texture acquisition without the need for polarized illumination.
Riviere et al.~\cite{disney_single_shot} build a similar lightweight system reintroducing polarization without active illumination, and modeling subsurface scattering.
This effort was refined to include global illumination and polarization modeling~\cite{xu2022improved}.
The proposed solutions deliver impressive visual results, but require expensive and difficult to use hardware.
We propose a solution for high-resolution facial texture reconstruction using commodity devices, such as smartphones.

\paragraph{Differentiable rendering.}
Recent progress in differentiable rendering~\cite{zeltner2021monte, vicini2021path, zhang2021path, Azinovic_2019_CVPR} has led to the development of mature frameworks~\cite{nimier2019mitsuba, jakob2022dr, Laine2020diffrast} and a number of methods that try to jointly estimate appearance and lighting~\cite{munkberg2022extracting}.
For an overview of differentiable rendering techniques, see~\cite{kato2020differentiable}.
Luan et al.~\cite{luan2021unified} use a co-located camera and light setup to reconstruct shape and material, relying on a differentiable renderer~\cite{zhang2021path} to produce unbiased gradients for shape estimation on an explicit mesh.
With the same co-located setup, Zhang et al.~\cite{zhang2022iron} improve results by using a hybrid volume radiance field and neural SDFs for the shape estimation.
While using a similar capture configuration to our work, the previous techniques focus on shape reconstruction, while we can lean on an accurate prior for the basis of our face shape.
Furthermore, by using polarization, we can properly decouple diffuse from specular textures.

Dib et al.~\cite{dib2021practical, dib2021towards, dib2022s2f2} propose the estimation of face skin textures by modelling the illumination with a virtual light stage, and using a differentiable ray tracer~\cite{Li:2018:DMC}.
The method fits a parametric face mask to the observed images and is able to handle self-shadowing, but complex lighting environments can have an impact on separation of lighting and material.
Wang et al.~\cite{wang2022sunstage} propose a capture setup with the sun as the main light source.
A FLAME~\cite{FLAME:SiggraphAsia2017} model is fit to the observed data, after which geometry and material are jointly refined using an analysis-by-synthesis approach.
As with other methods in uncontrolled lighting, separation of individual textures remains a challenge.

\paragraph{Deep learning-based approaches.}
A wide range of work proposes learning a neural network from large collections of high-quality light stage data, and subsequently applying the model to new data~\cite{meka2019deep, Saito_2017_CVPR, li2020dynamic, lattas2020avatarme, huynh2018mesoscopic, yamaguchi2018high, bi2021deeprelightable}.
Zhang et al.~\cite{zhang2021nlt} propose learning a neural light transport model from uv-space light and view direction information.
At test time, the model generalizes to novel views and lighting.
Several other works propose learning neural rendering models, either from single-view~\cite{Gafni_2021_CVPR,grassal2021neural,Zheng:CVPR:2022,Sengupta_2021_ICCV} or multi-view~\cite{sun2021nelf, sevastopolsky2020relightable} data, for a range of different applications.

\section{Method}

We propose a two-step analysis-by-synthesis approach for the estimation of high resolution face textures, as depicted in \Cref{fig:pipeline}.
The user captures two video sequences and a series of photographs of their face under linear-polarized point light illumination using a smartphone.
The first sequence has the polarization filters oriented in a perpendicular fashion, i.e., the filter covering the camera lens is perpendicular to the filter covering the smartphone's flashlight.
In accordance with existing literature, we denote this sequence as the cross-polarized sequence.
The second video sequence has parallel oriented filters and will be referred to as the parallel-polarized sequence.
We use structure-from-motion and multiview-stereo on all captured frames jointly to compute the camera alignment and reconstruct coarse geometry in form of a triangle mesh.
We then non-rigidly fit the FLAME model~\cite{FLAME:SiggraphAsia2017} to the scan and use it as our base geometry model.
This fitting helps us avoid noise from the multiview-stereo and provides a consistent UV-parameterization for all subjects.
Based on this geometry, we recover the diffuse albedo texture of the subject using the cross-polarized data and photo-metric optimization.
While keeping the diffuse albedo fixed, we estimate the remaining textures based on the parallel-polarized data.
Note that we reconstruct textures using only the photographs, as these capture more detail than the video frames.
For the geometry reconstruction, we use all captured data, as we found that this leads to more robust results compared to only using a small set of photographs.

\subsection{Capturing Polarized Data with a Smartphone}
\label{sec:data_capture}

\begin{figure}[t!]
    \begin{center}
        \includegraphics[width=0.94\linewidth]{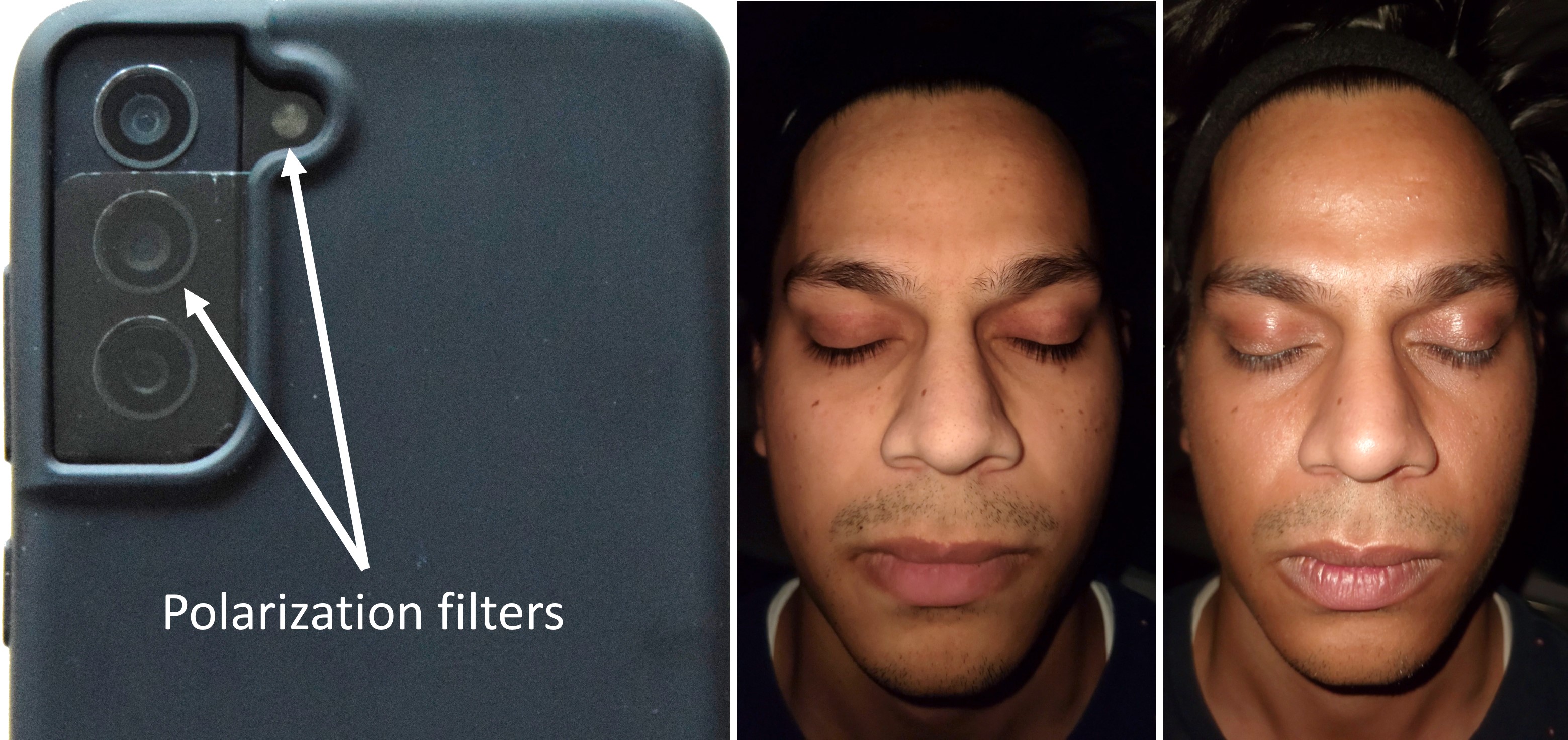}
    \end{center}
    \vspace{-0.5cm}
    \caption{
        Left to right: smartphone equipped with polarization filters, cross-polarized image (perpendicular filter orientation) and parallel-polarized image (parallel filter orientation).
    }
    \vspace{-0.35cm}
    \label{fig:capture_setup}
\end{figure}

We capture one cross-polarized and one parallel-polarized video sequence with a smartphone in a dark room, with the smartphone's flashlight as the only source of illumination.
Such a capture setup has the advantage of not requiring optimization of the scene lighting, leading to better separation of appearance and shading.
We assume that the flashlight is co-located with the camera lens and that its color is white.
We capture a color-checker under both filter orientations to color-calibrate both sequences.
This is important, since the filters introduce wavelength-dependent attenuation which tints the color of the light.
We use an affine color calibration scheme to compute the corresponding color correction matrix only once, and apply it to all subsequent sequences.
Furthermore, since an arbitrary smartphone's flashlight does not behave like an ideal point light (e.g., due to occlusion by the phone's cover along grazing directions), we pre-compute a per-pixel light attenuation map, that is multiplied with the final rendered images during optimization.
To this end, we put markers on a flat white surface and record a cross-polarized sequence of the surface.
We form an optimization problem with the unknowns being the surface's diffuse texture and the per-pixel light attenuation map.
The map is then kept fixed for all future face texture optimizations.
We refer to the supplemental material for more detail on this calibration step.
We ensure that all captures have consistent and fixed camera settings: focal length, exposure time and white balance.
We capture at 4K resolution and 30fps and select the sharpest frame from every 10-frames window, using variance of the Laplacian as the sharpness metric.
In addition to the video data, we capture a set of cross-polarized and a set of parallel-polarized photographs to obtain higher-quality data.
Since the flash is much brighter for photographs than for videos, we capture the photographs with shorter exposure and lower ISO to roughly match the brightness of the video frames.
The entire capture takes about five minutes.

\subsection{Geometry Reconstruction}
\label{sec:geometry_reconstruction}

\begin{figure}[t!]
    \begin{center}
        \includegraphics[width=0.93\linewidth]{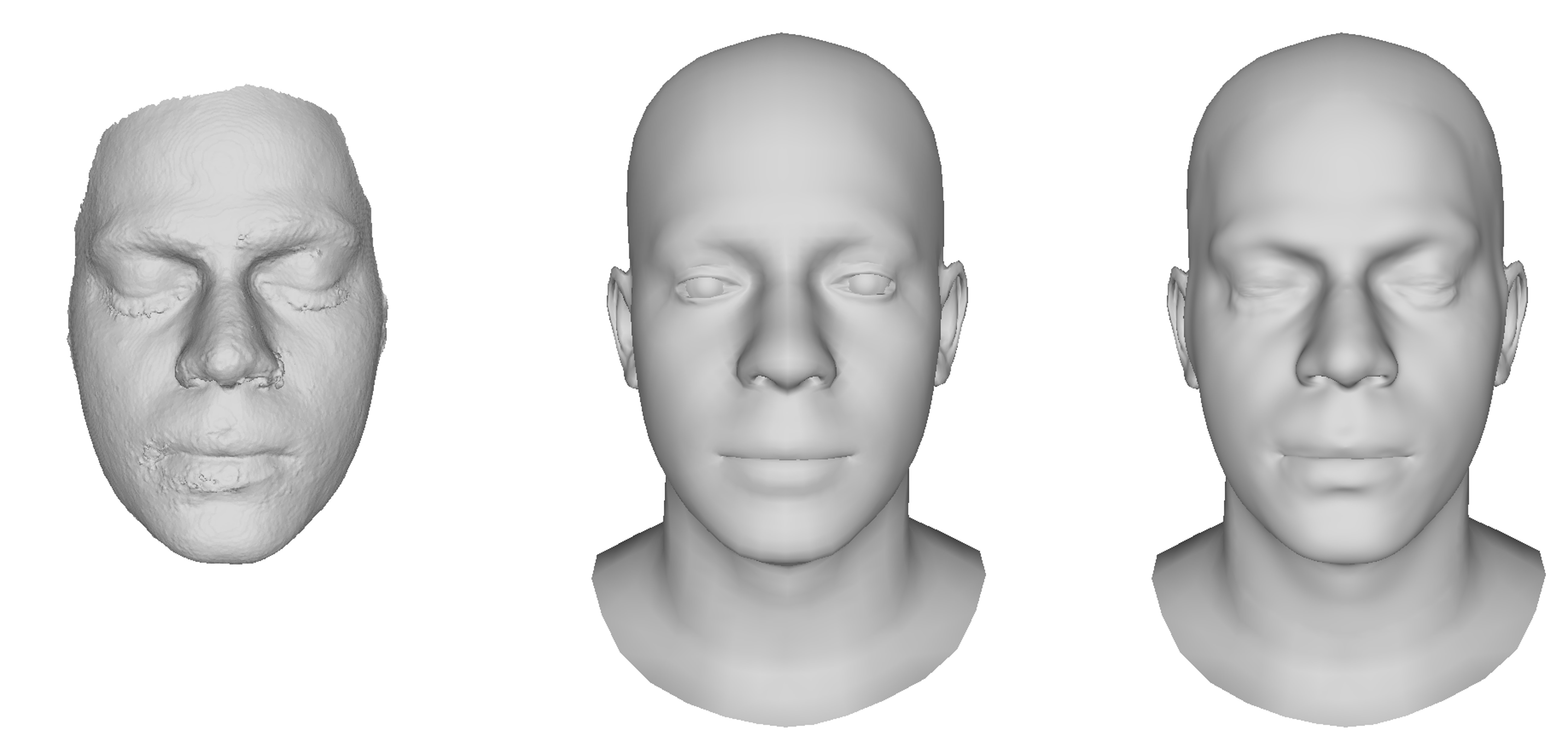}
    \end{center}
    \vspace{-0.5cm}
    \caption{
        Geometry reconstruction for subject from Figure~\ref{fig:capture_setup}. From left to right:
        reconstruction via structure from motion, fitted FLAME~\cite{FLAME:SiggraphAsia2017} mesh, ICP-based refinement of the mesh. 
    }
    \vspace{-0.2cm}
    \label{fig:geometry}
\end{figure}

We use Agisoft Metashape~\cite{metashape} on all frames jointly to obtain an initial mesh reconstruction.
We provide Metashape with face masks estimated by~\cite{zheng2021farl}, to make the reconstruction more robust to rigid motion of the head.
We then fit the FLAME model~\cite{FLAME:SiggraphAsia2017} to the scanned geometry, first by optimizing the shape parameters of the FLAME face space, and then by an ICP-based as-rigid-as-possible deformation approach (see \Cref{fig:geometry}).
For the non-rigid deformation, we subdivide the triangles of the face region, to obtain detailed geometry.
The resulting mesh is used as the base mesh for the subsequent texture optimizations.

\subsection{Rendering Equation \& BRDF}
\label{sec:bsdf_model}

We model the skin with a spatially-varying bidirectional reflectance distribution function (SV-BRDF).
Assuming a point light source $\textbf{l}$ in a dark environment, the rendering equation that defines the outgoing radiance $L_o(\textbf{x}, \omega)$, at point $\textbf{x}$ with normal direction $\textbf{n}^\top$ in direction $\omega$, has the following simplified form:
\begin{equation}
    L_o(\textbf{x}, \omega) = \frac{f(\textbf{x}, \omega) (\textbf{n}^\top \omega) L_i(\textbf{x}, \omega)}{\left| \textbf{x} - \textbf{l} \right|_2^2}.
\end{equation}
Here, we make use of the fact that the light direction aligns with the view direction, \ie, $\omega_i = \omega_o = \omega$.
The BRDF $f(\textbf{x}, \omega)$ has a diffuse component $f_d$, and a specular component $f_s$.
We use the Cook-Torrance~\cite{cook1982reflectance} BRDF for our specular term:
\begin{equation}
    f_s(\textbf{x}, \omega) = k_s(\textbf{x}) \frac{D(\omega, \textbf{n}^\top, \alpha)G(\textbf{n}, \omega)F(\textbf{n}, \omega)}{4 (\textbf{n}^\top \omega)(\textbf{n}^\top \omega)},
\end{equation}
with $k_s$ being the spatially-varying specular gain and $\alpha$ a global roughness blend factor for the Blinn-Phong distribution term $D$ of the 2-lobe mix ($D_{12}$ and $D_{48}$) suggested by~\cite{disney_single_shot}.
$G$ denotes the geometry term of the Cook-Torrance BRDF model.
We use Shlick's approximation~\cite{Schlick:1994} for the Fresnel term $F$:
\begin{equation}
    F(\textbf{n}, \omega) = F_0 + (1 - F_0) (1 - \textbf{n}^\top \omega)^5.
\end{equation}
To model the skin's diffuse response, we implement the BRDF model proposed by Ashikhmin and Shirley~\cite{ashikhmin2000anisotropic, ashikhmin2002anisotropic}, that accounts for the fact that a portion of the light has already scattered before penetrating the skin surface:
\begin{equation}
    f_d(\textbf{x}, \omega) = \frac{28 k_d(\textbf{x})}{23 \pi} (1 - F_0) (1 - (1 - \frac{\textbf{n}^\top \omega}{2})^5)^2,
\end{equation}
where $F_0 = 0.04$ is the reflectance of the skin at normal incidence.
Indirect light bouncing from the capture environment and on the captured face itself might have a significant contribution to pixel intensity at grazing angles, so we also add a Fresnel-modulated ambient term to our BRDF $f$:
\begin{equation}
    f_a(\textbf{x}, \omega) = k_a(\textbf{x}) (1 - (1 - F_0) (1 - (1 - \frac{\textbf{n}^\top \omega}{2})^5)^2),
\end{equation}
with an ambient map $k_a$ which is regularized to be smooth via a total variation loss and close to zero.

Note that using a diffuse scattering model for the optimization is compatible with state-of-the-art physically-based subsurface scattering skin shading~\cite{wrenninge2017path, chiang2016practical}, as shown in \Cref{fig:teaser}.
Production-ready subsurface scattering models typically include an albedo inversion stage, which takes a diffuse albedo as input, and converts it to extinction coefficients for the volume rendering random walk.

\subsection{Optimization}
\label{sec:optimization}

The objective of the photometric optimization step is to minimize the difference between rendered images $\hat{I}$ and color-corrected target images $I$:
\begin{equation}
    \mathcal{L}(\hat{I}, I) = \left| W\cdot \left( \hat{I} - I \right) \right|,
\end{equation}
with $\hat{I}=\mathcal{M} \cdot L_o$, where $\mathcal{M}$ is the pre-computed light attenuation map, that accounts for uneven light distribution in different directions.
We apply a per-pixel loss weight $W$ based on the respective mip level and the angle between viewing direction and normal $\textbf{n}^\top \omega$ to improve sharpness.
Specifically, to ensure that distant or grazing angle observations do not blur the resulting textures, for each pixel that is projected from the target image to texture space, we calculate which mip level $l$ would need to be looked up in classical forward rendering.
$W$ is set to $(\textbf{n}^\top \omega)(1 - l)$ if the pixel corresponds to a mip level below 1, and zero otherwise.
We optimize $\mathcal{L}(\hat{I}, I)$ in two steps, using a coarse-to-fine optimization strategy in each.
In the first step, we only use the cross-polarized images to optimize the spatially-varying diffuse albedo texture $k_d(\textbf{x})$ and an initial tangent-space normal map $n(\textbf{x})$, while assuming $f_s(\cdot) = 0$ for the specular term.
In the second step, we fix the diffuse texture and optimize for specular gain $k_s(\textbf{x})$, specular roughness $\alpha$, and the final normal map $n(\textbf{x})$.
To account for potentially different light attenuation in the cross and parallel-polarized filter settings, we also optimize per-channel scaling factors for the diffuse texture.
The optimization is performed entirely in texture space.
In each step, we employ a four-level coarse-to-fine optimization strategy, starting with a texture resolution of $512\times512$, and increasing the size by a factor of two after convergence of each level, up to the final resolution of $4096\times4096$.
We implement our optimization framework in PyTorch, using nvdiffrast~\cite{Laine2020diffrast} as our differentiable renderer.
We optimize on batches of 4 images, using Adam with an initial learning rate $lr_0 = 10^{-3}$ for all parameters at the beginning of every coarse-to-fine step, and updating it to $lr = lr_0 \cdot 10^{-0.001t}$ in every iteration $t$.
We scale the FLAME mesh to unit size and set the light intensity to 10.
The total optimization time is about 90 minutes.
\vspace{-0.1cm}
\section{Results}
\label{sec:results}
\vspace{-0.1cm}

\begin{figure*}[!htb]
    \begin{center}
    \includegraphics[width=0.98\linewidth]{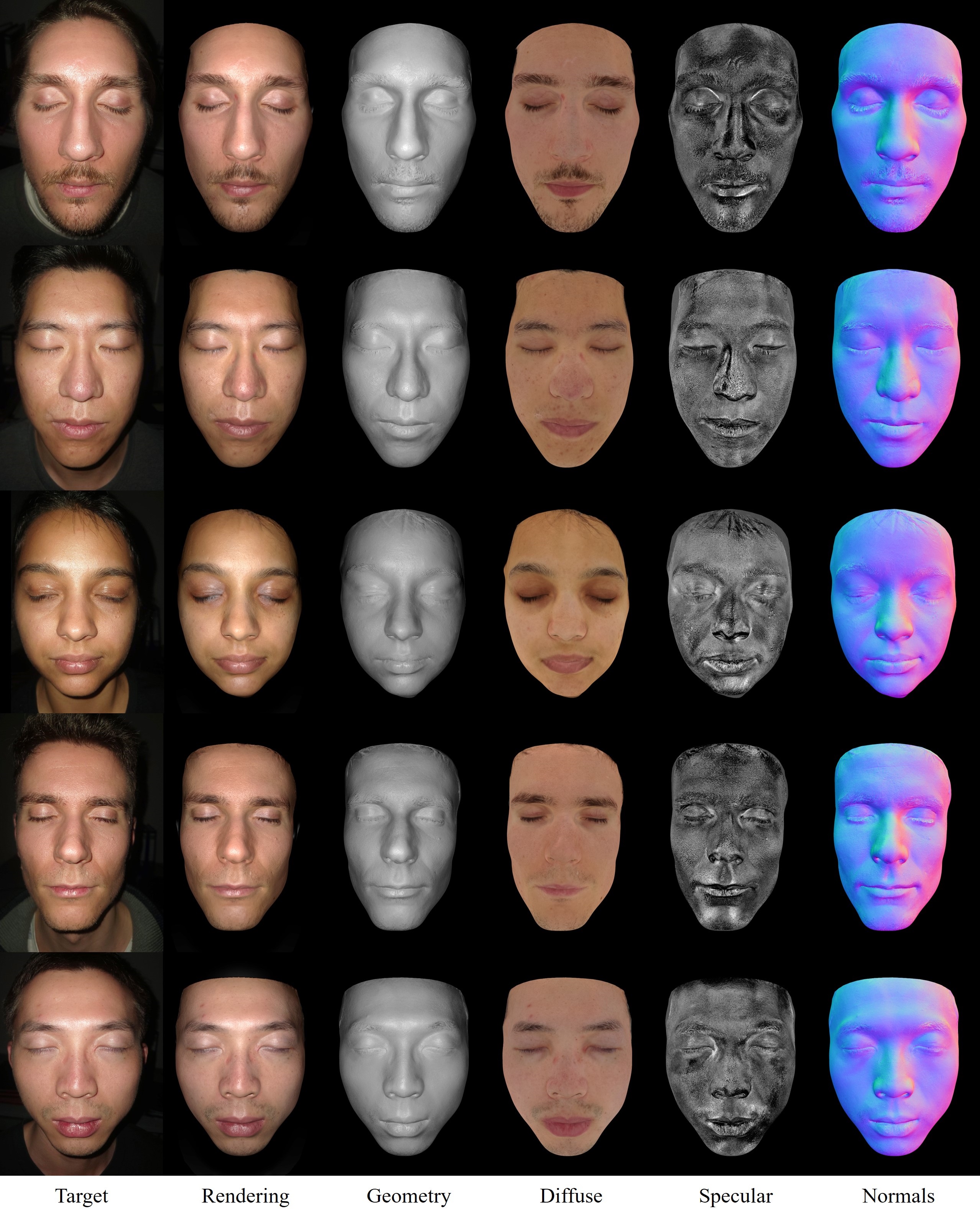}
    \end{center}
    \vspace{-0.5cm}
   \caption{We show skin texture reconstructions of several actors of different skin type. The rendered images closely match the reference target images, and we achieve good separation of diffuse and specular textures.
   }
\label{fig:results}
\end{figure*}

\begin{figure*}[!htb]
\begin{center}
\includegraphics[width=0.95\linewidth]{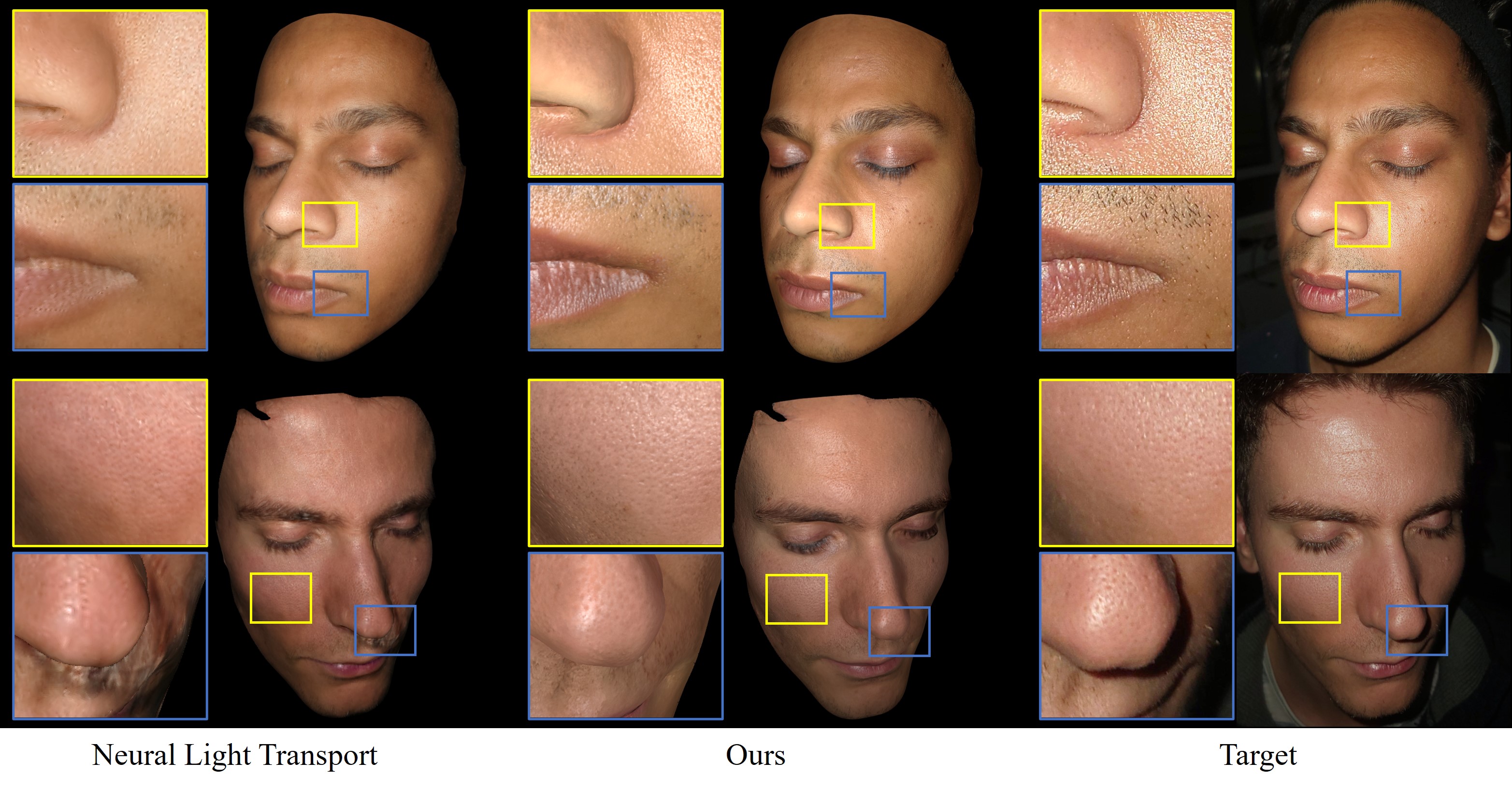}
\vspace{-0.6cm}
\end{center}
    \caption{We evaluate on a validation frame from a novel viewpoint and with novel lighting that was held out during the optimization. As visible in the crop regions, our method is able to synthesize sharper texture details and specular highlights compared to NLT~\cite{zhang2021nlt}.
    }
    \label{fig:relighting}
\vspace{-0.2cm}
\end{figure*}

In this section, we present texture reconstruction and rendering results on several subjects.
Figure~\ref{fig:results} shows the texture reconstruction on several actors of different ethnicity.
Our method is able to reconstruct pore-level detail in the diffuse, specular and normal maps.
Further, we evaluate the quality of our reconstructed textures by rendering the mesh from novel views and under novel illumination.
Figure~\ref{fig:relighting} shows that our method faithfully reconstructs the skin's appearance under novel views and lighting.

\begin{figure}[!htb]
\begin{center}
\includegraphics[width=0.975\linewidth]{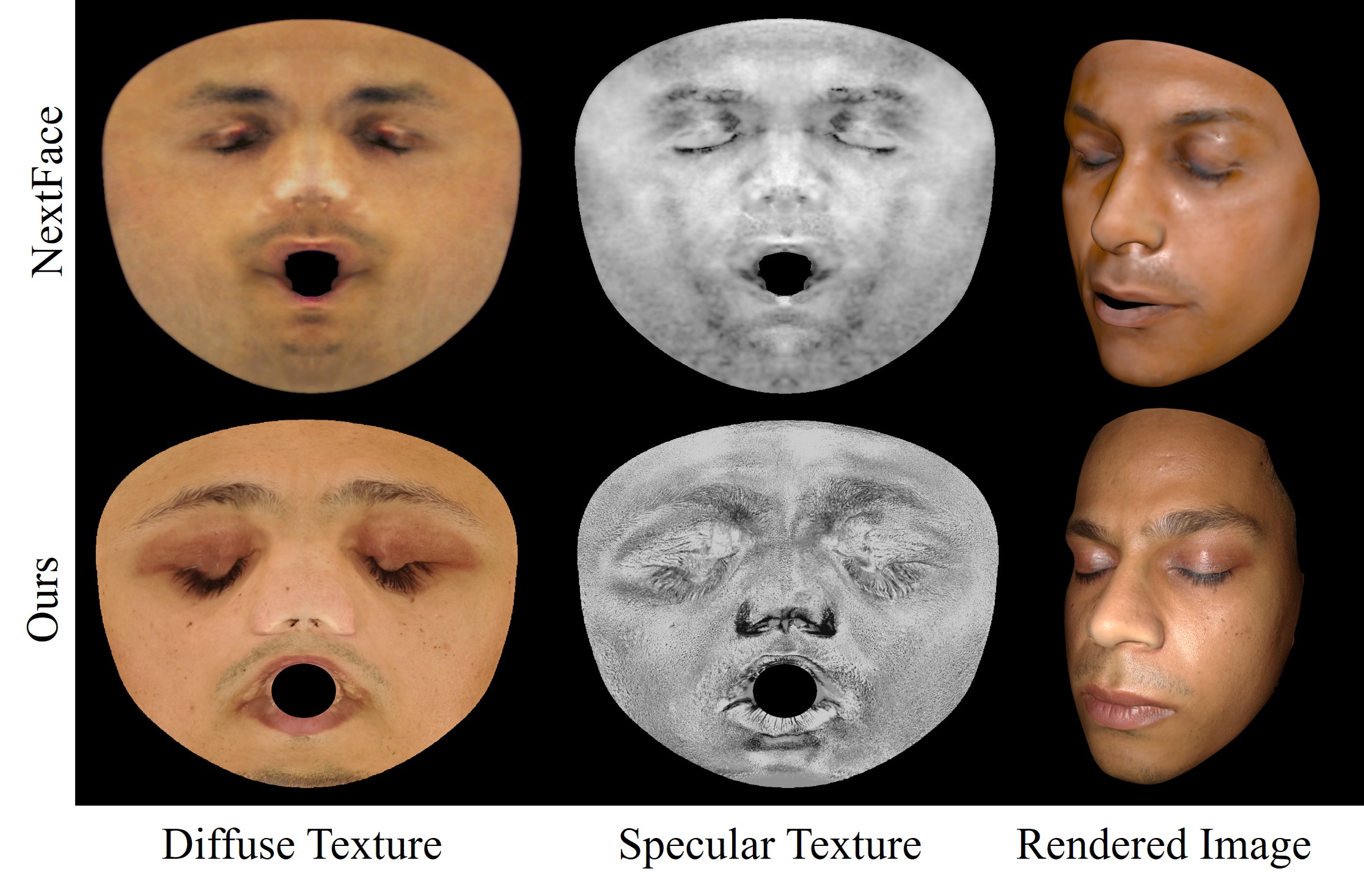}
\end{center}
\vspace{-0.6cm}
   \caption{
   Comparison to NextFace~\cite{dib2022s2f2} in terms of reconstructed appearance textures.
   }
\label{fig:textures}
\vspace{-0.4cm}
\end{figure}

\paragraph{Comparison to state of the art.}
We perform both a qualitative and quantitative evaluation of our method and compare to state-of-the-art methods for relighting and texture reconstruction.
During optimization, we hold out a validation frame on which we compute image metrics.

\begin{table}
    \centering
	\resizebox{0.8\linewidth}{!}{
    \begin{tabular}{lccc}
        \toprule
        \textbf{Method}  & \textbf{PSNR} $\uparrow$  & \textbf{SSIM} $\uparrow$ & \textbf{LPIPS} $\downarrow$ \\
        \midrule
        NLT~\cite{zhang2021nlt}
        & 31.51                     & \textbf{0.96}                     & 0.11  \\
        NextFace~\cite{dib2022s2f2}        
        & 22.85                     & 0.89                     & 0.31  \\
        \midrule
        Ours             
        & \textbf{32.37}            & \textbf{0.96}           & \textbf{0.10} \\
        \bottomrule
    \end{tabular}
    }
    \caption{We compare our method to NLT and NextFace on validation frames over 10 different subjects.}
    \label{tab:quantitative}
    \vspace{-0.3cm}
\end{table}

\emph{Neural Light Transport.}
Neural Light Transport~\cite{zhang2021nlt} is a deep learning-based method that takes as input pre-computed diffuse base, light-cosine and view-cosine uv-space maps.
The diffuse base is computed as the average of all observations.
The cosine maps contain per-texel cosines of the angles between the normal vector and the light or view vector.
Based on these inputs, as well as nearest neighbor observations, a neural network learns to predict the final shaded image.
Since the method does not take light intensity and falloff into account, we optimize the rendered validation image's brightness to match the target as closely as possible, before computing the rendering error.

\emph{NextFace.}
NextFace~\cite{dib2021practical, dib2021towards, dib2022s2f2} first fits a morphable face model to the input frames, then estimates the face shape, pose, lighting, statistical diffuse and specular albedos by minimizing a photo-consistency loss between the target image and a ray traced estimate.
In a final step, the statistical albedos are refined on a per-texel basis.
We conducted several experiments with different illumination conditions and number of frames, including an experiment on our data for which we replaced the spherical harmonics lighting representation with a small area light, modelling our flashlight.

As shown in \Cref{tab:quantitative}, our approach achieves favorable image metrics.
\Cref{fig:relighting} compares our method to NLT on novel lighting and viewpoint.
NLT closely matches the target by using nearby camera views, but specular highlights are often blurry, and the low number of training views results in the model producing artifacts in shadowed areas.
We obtained the best NextFace results in an experiment with uniform illumination using three frames that cover the whole face region.
As shown in \Cref{fig:textures}, inaccuracies in the face model fitting lead to somewhat blurry textures.
This issue is exacerbated by adding more frames.
Using fewer frames degraded the separation of the diffuse and specular textures.
Our method is able to overcome these issues by accurately fitting a geometric model to the input data and by using polarization to separate the individual textures.

\begin{figure}[t!]
\begin{center}
\vspace{-0.1cm}
\includegraphics[width=0.975\linewidth]{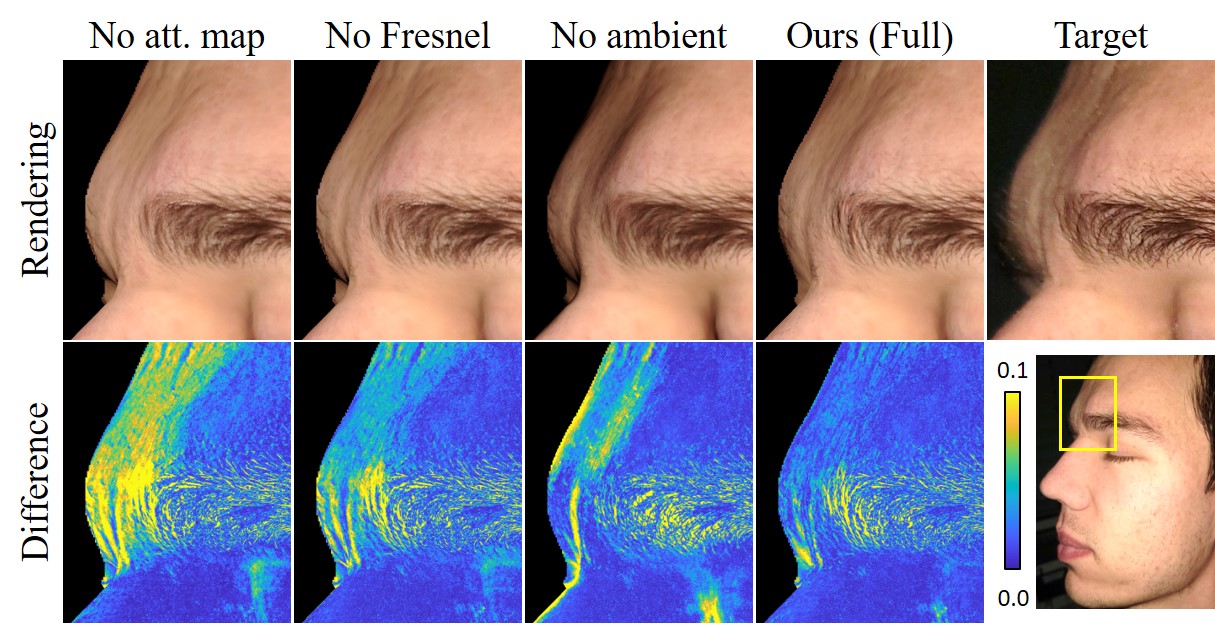}
\vspace{-0.6cm}
\end{center}
   \caption{Ignoring the angle-dependent flashlight attenuation, the Fresnel effect, or the ambient light leads to an incorrect reconstruction, that can no longer reproduce the shading from all views. We account for these effects to closely match the target data.}
\label{fig:ablations}
\end{figure}

\begin{figure}[t!]
\begin{center}
\includegraphics[width=0.975\linewidth]{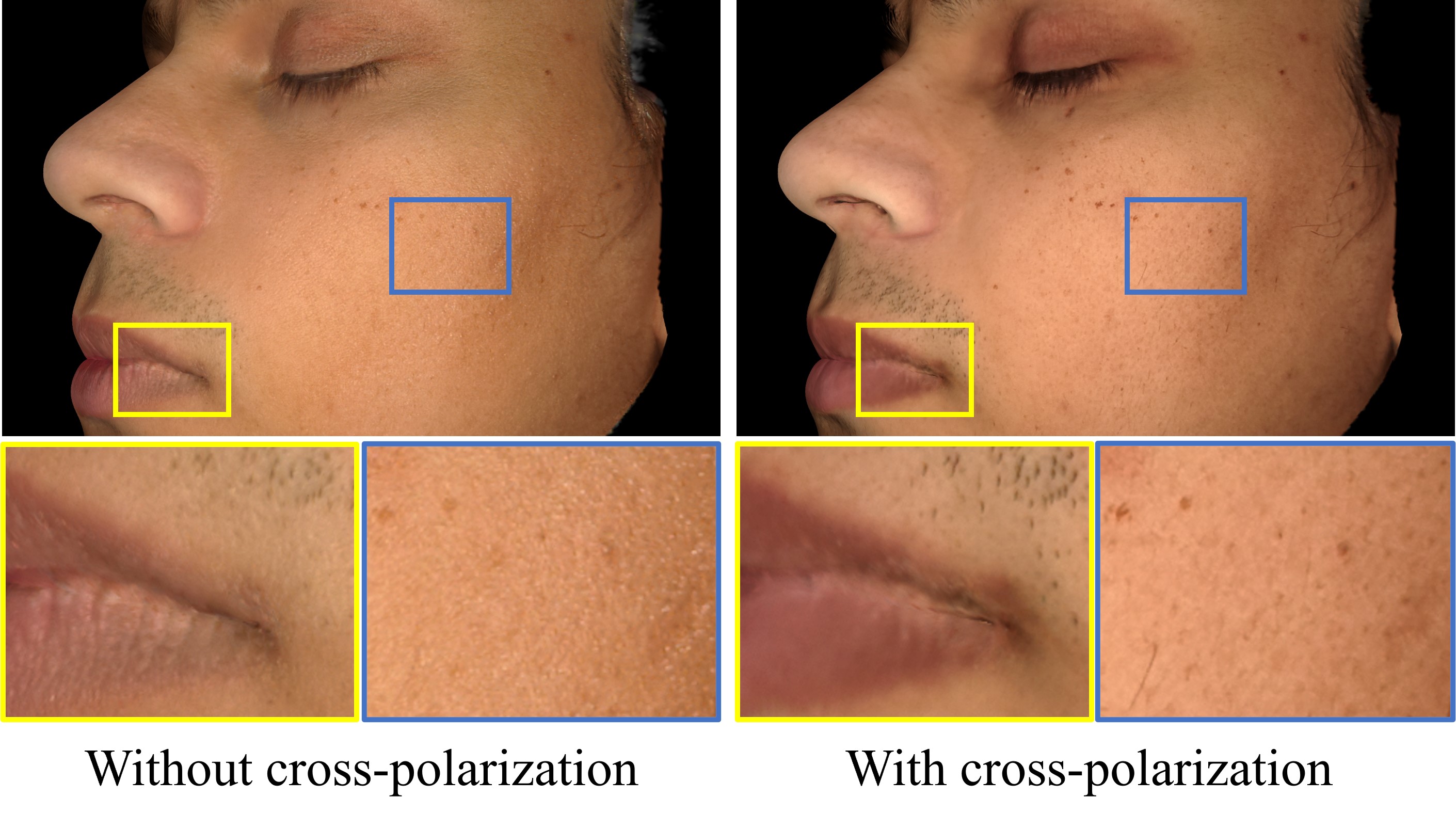}
\vspace{-0.7cm}
\end{center}
   \caption{We compare joint optimization of all textures to our full approach on a purely diffuse render. Optimizing jointly leaks specular and normal map information into the diffuse texture.}
\label{fig:no_polarization}
\vspace{-0.2cm}
\end{figure}

\paragraph{Ablation Studies.}
We conduct ablation studies to justify our choice of capture setup and training parameters.
In Figure~\ref{fig:ablations}, we show that accounting for the direction-dependent light attenuation of a smartphone's flashlight leads to an overall lower error in the re-rendered images.
In the same figure, we also show the importance of accounting for the Fresnel effect when reconstructing the diffuse texture.
A purely Lambertian BRDF will not be able to model the skin's diffuse response at all angles.
In \Cref{fig:no_polarization}, we show that optimizing textures without cross-polarization will leak specular information into the diffuse texture.

\begin{figure}[t!]
\begin{center}
\includegraphics[width=0.975\linewidth]{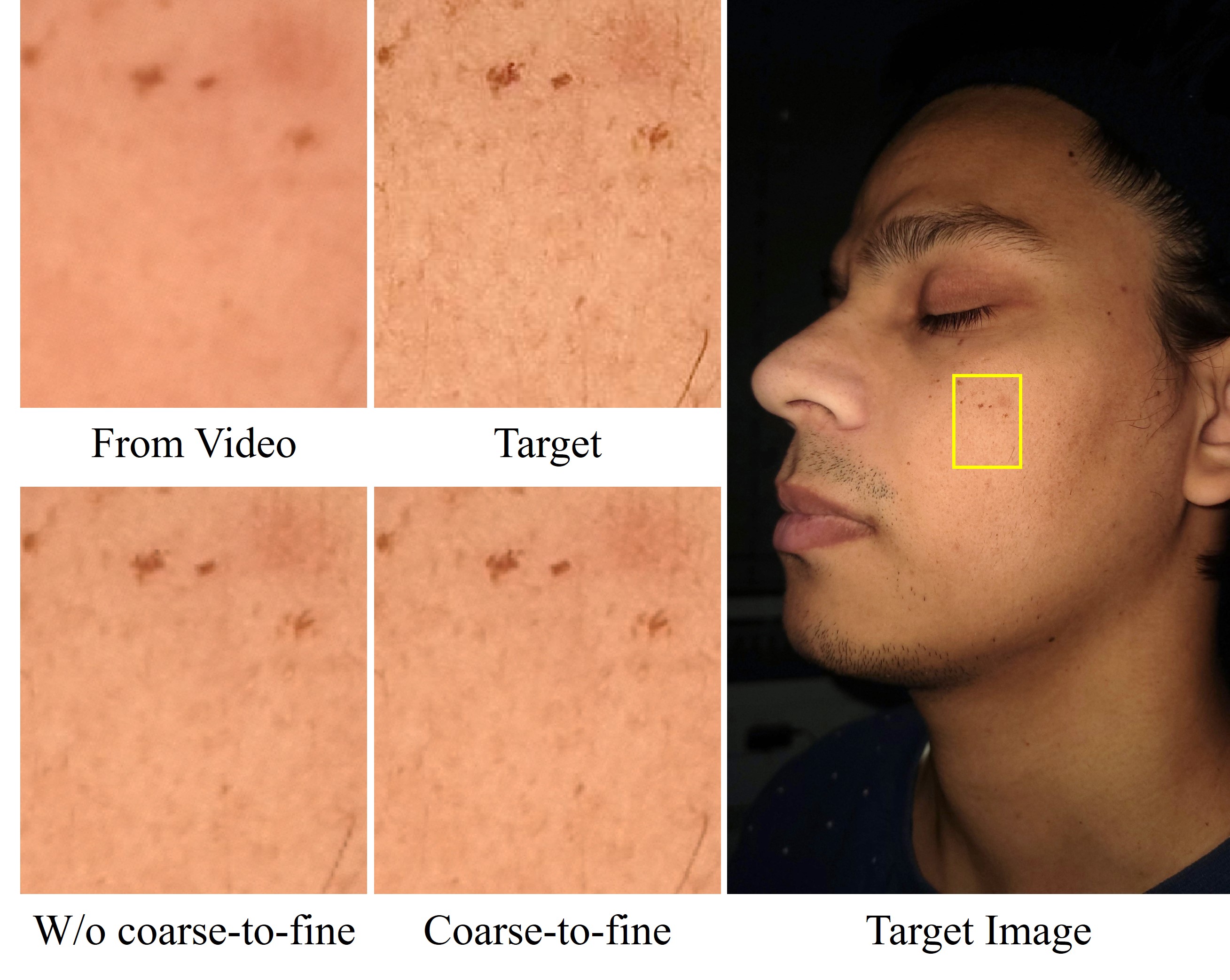}
\vspace{-0.7cm}
\end{center}
   \caption{To increase sharpness, we optimize from photographs, instead of video frames. Using only pixels of the appropriate mip level in a coarse-to-fine approach further enhances results.}
\label{fig:mip}
\vspace{-0.2cm}
\end{figure}

\paragraph{Coarse-to-fine optimization and mipmapping.}
Pixels of the target images have different footprints in uv-space, depending on distance and angle between camera and surface.
Weighting the loss of each pixel equally leads to blur in the reconstruction.
Optimizing coarse-to-fine, where at each resolution we use only pixels with the corresponding uv-space footprint, helps us reconstruct additional detail in the textures.
Figure~\ref{fig:mip} shows a comparison between our full approach and a direct optimization of the highest resolution texture.
We additionally show the decrease in quality when optimizing only on video frames (w/o photographs).

\paragraph{Runtime and memory consumption.}
Including $1$ hour spent on MVS, our method needs about $2.5$h to reconstruct a face. 
Photo-metric skin texture reconstruction takes about 90 minutes on an Nvidia RTX A6000.
We reconstruct facial geometry with Metashape using an average of 420 video frames and 70 photographs.
At a texture resolution of $4096\times4096$ and target image resolution of $3840\times2160$, the photo-metric optimization requires $30$GB of GPU memory.
In comparison, NLT takes about 10h and NextFace about 6h given the same number of frames.

\paragraph{Discussion \& Limitations.}
Our method reconstructs high-quality face textures with a low-cost capture routine.
However, it is restricted to static expressions, i.e., it does not handle dynamically changing face geometry and textures.
An avenue for future research is the reconstruction of dynamic expressions by fitting a parametric model with consistent mesh topology to each frame, and optimizing over the entire non-rigid sequence.
Our method does not explicitly handle global illumination.
A differentiable path tracer could potentially improve results in the concavities of the eye region.
As we assume a static face with closed mouth and closed eyes, we only recover the skin area of a face.
Eyes, mouth interior and hair are a subject of future work.

\section{Conclusion}

We have presented a practical and inexpensive method of capturing high-resolution textures of a person's face by coupling commodity smartphones and polarization foils.
The co-location of the camera lens and light source allows us to reduce the problem complexity and separate material from shading information.
As a result, we obtain high-resolution textures of the skin area of the human face.
We believe that polarization is a powerful tool for material recovery in the real world, and future smartphones could benefit from including filters directly in the hardware.
Overall, we believe that our work is a stepping stone towards democratizing the creation of digital human face assets by making it more accessible to smaller production studios or individual users.

\vspace{-0.2cm}
\section*{Acknowledgements}
\vspace{-0.2cm}
{
Dejan Azinovi{\'c}'s contribution was supported by the ERC Starting Grant Scan2CAD (804724), the German Research Foundation (DFG) Grant ``Making Machine Learning on Static and Dynamic 3D Data Practical'', and the German Research Foundation (DFG) Research Unit ``Learning and Simulation in Visual Computing''. We would also like to thank Angela Dai for the video voice over and Simon Giebenhain for help with the FLAME fitting.
}

{\small
\bibliographystyle{ieee_fullname}
\bibliography{egbib}
}

\begin{appendix}
\clearpage
\newpage
\section*{APPENDIX}

In this appendix, we describe in more detail the pre-processing steps that are necessary to run our method.
Specifically, in Section~\ref{sec:calibration} we explain in detail how to calibrate the camera light and sensor and in Section~\ref{sec:geometry_estimation} we give more detail on fitting the FLAME mesh to the Structure-from-Motion scan.
In addition to that, we discuss differences to prior work in Section~\ref{sec:prior_work}.

\section{Calibration}
\label{sec:calibration}

\begin{figure}[t!]
    \begin{center}
        \includegraphics[width=\linewidth]{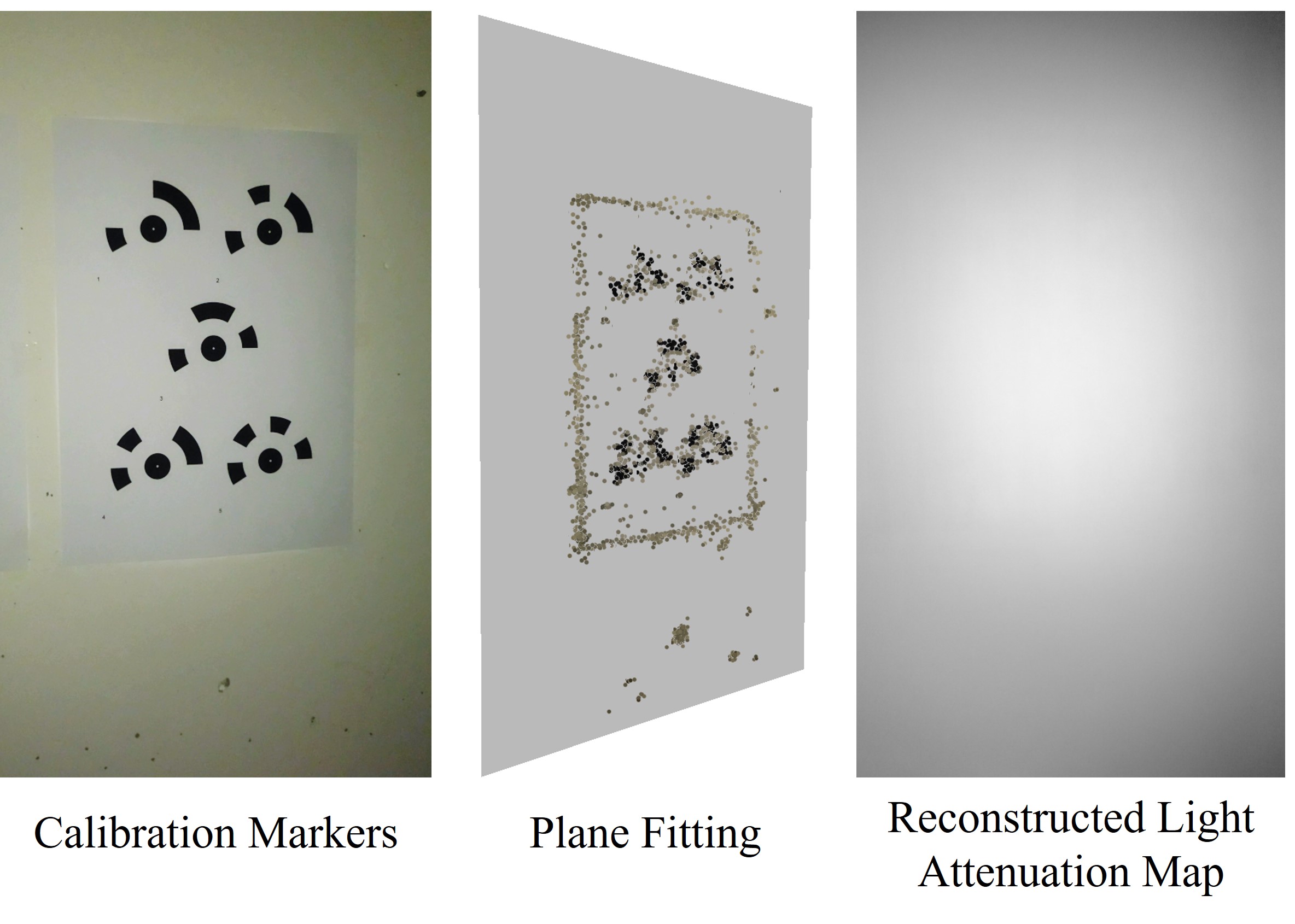}
    \end{center}
    \vspace{-0.4cm}
    \caption{
        To calibrate the light of the smartphone, we record a cross-polarized sequence of a white planar surface with markers for tracking.
        We fit a UV-parameterized plane to the data and optimize for a light attenuation map which we use for all experiments.
    }
    \vspace{0.2cm}
    \label{fig:light_attenuation}
\end{figure}

To use a smartphone as a tool to capture high-quality textures of human faces, we apply a calibration step related to the flashlight and camera sensor.
Specifically, we compute a light attenuation map to take into account vignetting effects and the fact that the flashlight is not an ideal point light source, and we color-calibrate the cross-polarized and parallel-polarized images.
\paragraph{Light attenuation map.}
In the general case, a smartphone's flashlight does not behave like an ideal point light.
We observed a significant decrease of light intensity towards grazing angles.
To account for this effect, we compute a per-pixel attenuation map that we multiply with our rendered images to match the observations.
To this end, we put calibration markers on a white wall and recorded a cross-polarized sequence (see \Cref{fig:light_attenuation}).
The markers allow us to estimate camera poses for the sequence and provide us a sparse point cloud to which we fit a plane.
Finally, we pose an optimization problem:
\begin{equation}
    \argmin_{\mathcal{M}, k_d} \left| \left( \hat{I} - I \right) \right|,
\end{equation}
with $\hat{I}=\mathcal{M} \cdot L_o$, where $\mathcal{M}$ is the light attenuation map, and $k_d$ the diffuse texture.
Once optimized, we keep $\mathcal{M}$ fixed for all subsequent face texture optimizations.

\begin{figure}[t!]
    \begin{center}
        \includegraphics[width=\linewidth]{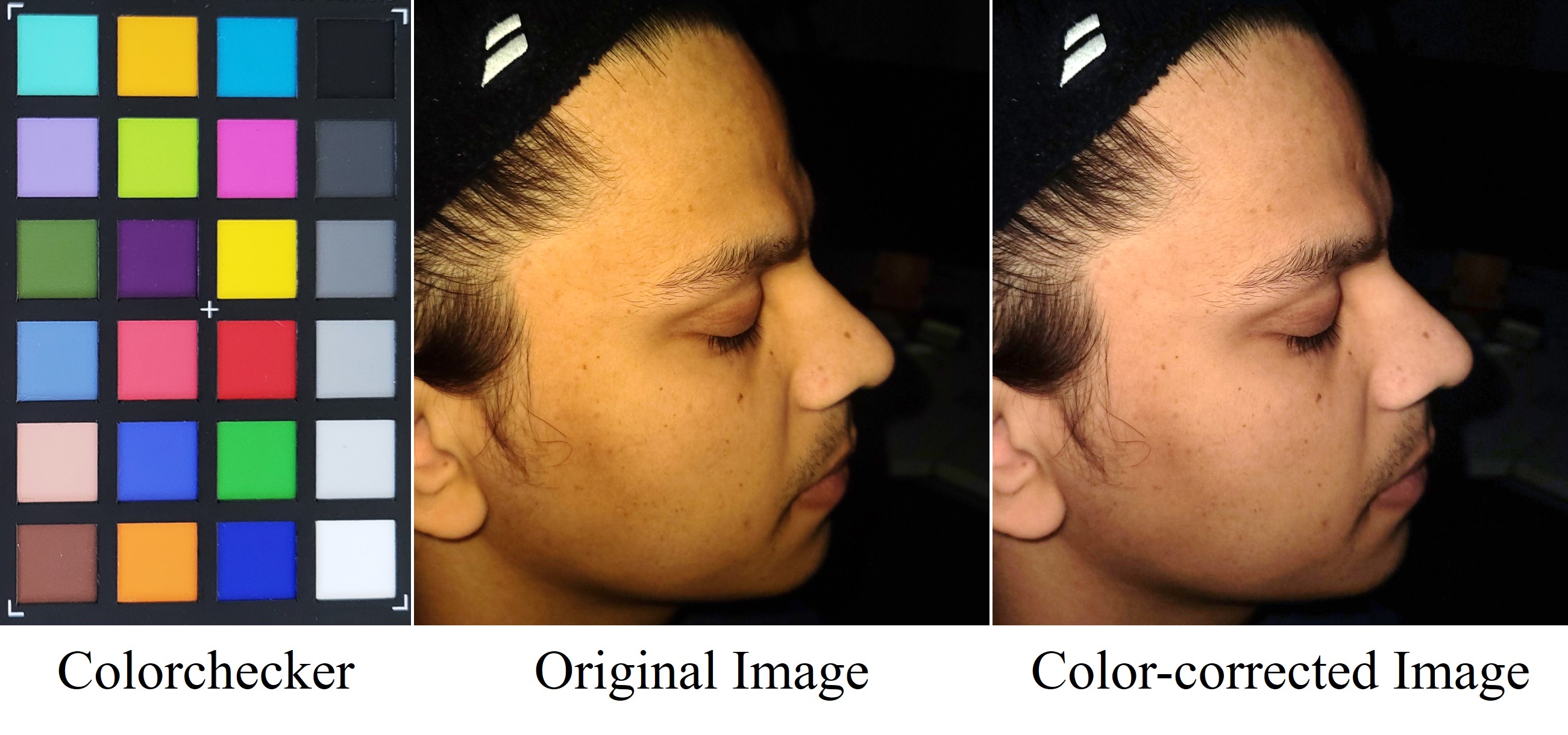}
    \end{center}
    \vspace{-0.4cm}
    \caption{
        We found that the polarization filters introduce a color shift depending on the polarization direction.
        To this end, we perform a color calibration with a Macbeth colorchecker board which we capture in both scenarios (cross-, and parallel-polarized).
        We use an affine color correction to match both captures, and apply this transformation to recordings of all subjects.
    }
    \vspace{-0.25cm}
    \label{fig:color_correction}
\end{figure}

\begin{figure}[t!]
    \begin{center}
        \includegraphics[width=\linewidth]{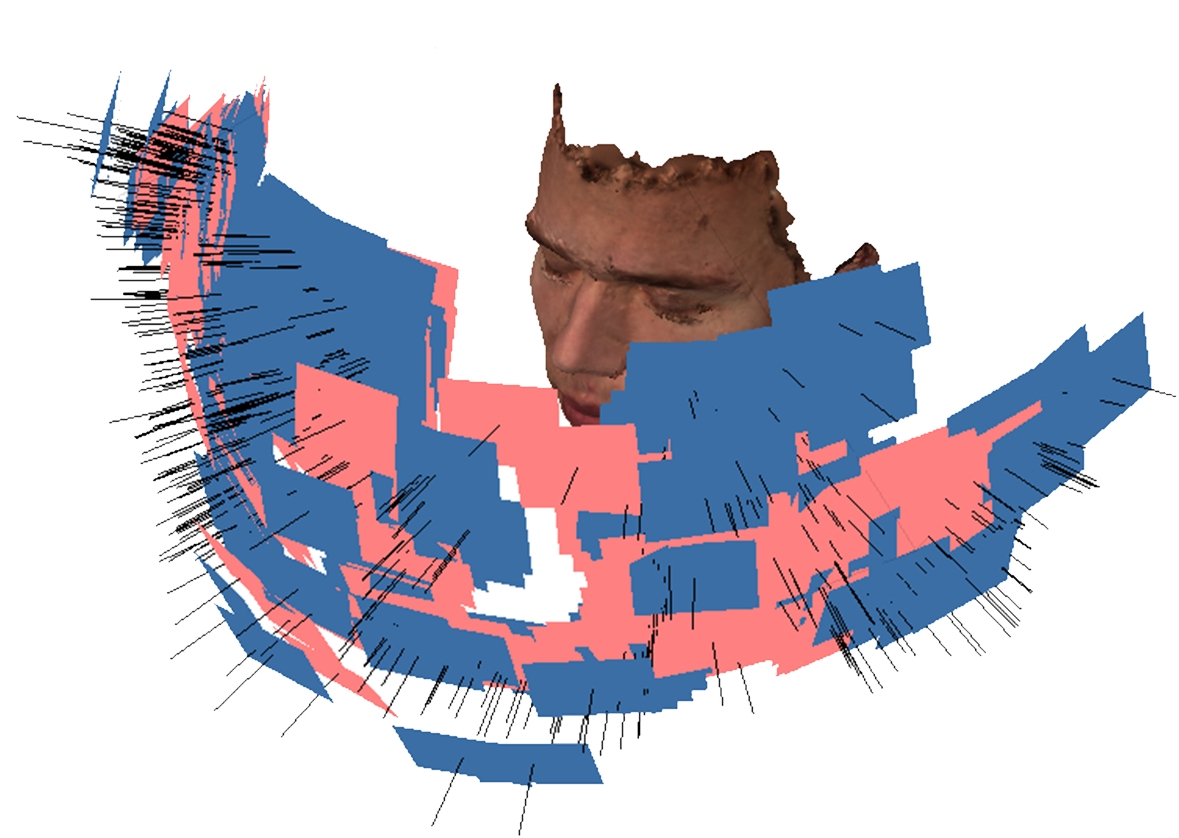}
    \end{center}
    \vspace{-0.4cm}
    \caption{
        Distribution of cross-polarized (red) and parallel-polarized (blue) views.
    }
    \label{fig:camera_visualization}
\end{figure}

\paragraph{Color correction.}
We color-calibrate both the cross-polarized images and the parallel-polarized images using pre-recorded images of a Macbeth colorchecker board.
We compute an affine color transformation matrix to match these calibration images to a reference color chart.
This calibration step is done once for the smartphone and then used for all recorded sequences.
The effect of this calibration step is shown in \Cref{fig:color_correction}.

\paragraph{Camera settings.}
We record our data using a Samsung Galaxy S21 FE 5G.
For the video sequences, we use an ISO of 800 and exposure time of 1/60s.
The photographs were shot with an ISO of 200 and exposure time of 1/90s.
The smartphone's white balance was set to 4900K.

\section{Geometry Estimation}
\label{sec:geometry_estimation}
To estimate the geometry of a subject, we use the Structure-from-Motion method from MetaShape~\cite{metashape} on the captured data (see Figure~\ref{fig:camera_visualization} for a camera pose visualization).
The resulting geometry is noisy and might contain holes, so we fit a 3DMM-based face model to the reconstruction.
Specifically, we use PIPNet~\cite{JLS21} to detect landmarks on a front-facing image of the face.
These are then projected to 3D using the known camera extrinsic and intrinsic matrices.
Using Procrustes's algorithm, we get a coarse alignment between the FLAME face model~\cite{FLAME:SiggraphAsia2017} and the 3D landmarks.
We further improve the alignment by optimizing for both a rigid transform between FLAME and nearby scan vertices, as well as the FLAME shape vector to non-rigidly fit the scan.
The resulting mesh is subdivided in the face region by a factor of 16, and the eyes are removed from the mesh.
Finally, we employ an As-Rigid-As-Possible (ARAP)~\cite{sorkine2007arap} non-rigid deformation strategy to refine the face mesh, to better align with the reconstruction of MetaShape.

\section{Comparison to Prior Work}
\label{sec:prior_work}
In this section, we explain in more detail the differences between our proposed method and results, and some of the existing solutions for light stage data to which we could not compare directly. Furthermore, we discuss potential benefits of capture setups with independent view and light directions.

\paragraph{MoRF}~\cite{morf} is a generative model trained on a high-quality image database with polarization-based separation of diffuse and specular reflectance.
It can generate a volumetric representation of a face based on latent ID codes, which can be optimized to fit new subjects.
The database itself is created using the capture setup from~\cite{disney_single_shot}.
Images of a subject can be rendered by first feeding the subject-specific ID code into a deformation and a canonical MLP.
The canonical MLP is composed of a density, diffuse and specular branch, and the output of these branches is used in a volumetric rendering formulation, similar to~\cite{mildenhall2020nerf}, to render the final image.
This is in contrast to our approach, which uses a triangle mesh to represent geometry, and which defines the SVBRDF on the surface of the mesh.
The major advantage of MoRF is the fewer number of images it requires at test time and better facial hair and eye handling.
This is, however, offset by its limited performance in accurately fitting to faces of new subjects.
Furthermore, the material is not separated from lighting and the results are over-smoothed due to the low-order spherical harmonics lighting approximation.

\paragraph{Deep Relightable Appearance Models for Animatable Faces}~\cite{bi2021deeprelightable} proposes a conditional variational auto-encoder (CVAE) architecture to predict mesh vertices, a corresponding texture warp field and light-dependent textures.
A late-conditioned model is first trained on light stage OLAT (one light at a time) data to predict a lit texture map of a subject's face from its average texture (nearest fully-lit frame averaged across all cameras) and an initial estimate of the mesh vertices (provided by an off-the-shelf face tracker).
This model has good generalization ability, but is not suitable for real-time rendering.
Making use of the good generalization ability of the trained model, a large dataset of synthetic images is generated and used to train an early-conditioned model which can render faces under complex lighting in real-time.
The biggest advantage compared to our approach is the capture and rendering of dynamic sequences.
Some of the drawbacks include the necessity of a light stage capture setup and the long training time.
Futhermore, the model does not separate lighting from material, so its output can not be used in a standard rendering pipeline, or for the creation of virtual assets. 

\paragraph{Near‐Instant Capture of High‐Resolution Facial Geometry and Reflectance}~\cite{fyffe2016near} performs multi-view color-space analysis to separate diffuse from specular reflectance.
Photometric estimation of specular normals further refines geometry compared to the reconstructed base mesh.
Similar to our method, and in contrast to the previously described deep learning-based methods, the output is a set of textures that can used in a standard rendering pipeline to render photo-realistic images of a person's face.
The carefully calibrated high-cost capture setup, consisting of 24 DSLR cameras, enables reconstruction of fine-scale detail and cannot be matched by current smartphone camera technology.
Nevertheless, we see potential benefit of our method's flexibility to capture specular highlights from arbitrary viewpoints, compared to a predetermined set of fixed viewpoints.
Another drawback is the necessity of a manual cleanup of the reconstructed multi-view stereo mesh, which is avoided by our method's automated FLAME fitting.

\medskip

Several prior works~\cite{lattaspractical,macek2022realtimerelighting,Sengupta_2021_ICCV} on face reconstruction and relighting use a capture setup, in which the light direction is independent from the view direction.
While we see potential benefit for convergence speed from the additional constraints provided by such capture setups, given multiple views, our co-located data also provides enough constraints for successful convergence.
The shadowing-masking term $G$ is the only term that is directly linked to both the view and light vector.
However, by reciprocity of the BRDF, the dependence on view and light direction is the same.
Instead of having independent view and light vectors, we found it more important to have a good distribution of the angles between surface normal and view (or light) vector to recover a complete specular and normal map.
This is in contrast to \cite{lattaspractical} and \cite{Sengupta_2021_ICCV} where both camera and light are mostly front-facing.
\end{appendix}

\end{document}